\documentclass{article}

\PassOptionsToPackage{numbers, compress}{natbib}

\usepackage[preprint]{neurips_2025}

\usepackage[utf8]{inputenc} 
\usepackage[T1]{fontenc}    
\usepackage{hyperref}       
\usepackage{url}            
\usepackage{booktabs}       
\usepackage{amsfonts}       
\usepackage{nicefrac}       
\usepackage{microtype}      
\usepackage{xcolor}         
\usepackage{graphicx}
\usepackage{capt-of}
\usepackage{adjustbox}
\usepackage{amsmath}
\usepackage{caption}
\usepackage{multirow}

\title{CompAlign: Improving Compositional Text-to-Image Generation with a Complex Benchmark and Fine-Grained Feedback}

\author{%
  Yixin Wan \quad \textmd{and} \quad Kai-Wei Chang \\
  Department of Computer Science\\
  University of California, Los Angeles\\
  \texttt{\{elaine1wan, kwchang\}@cs.ucla.edu} \\
}

\begin{document}

\maketitle

\begin{abstract}
State-of-the-art Text-to-Image models (e.g., DALL-E) are capable of generating high-resolution images given textual prompts. 
However, these models still struggle with accurately depicting compositional scenes that specify multiple objects, attributes, and spatial relations.
We present \textbf{CompAlign}, a challenging benchmark with an emphasis on assessing the depiction of 3D-spatial relationships, for evaluating and improving models on compositional image generation. 
CompAlign consists of \textbf{900 complex multi-subject image generation prompts} that combine numerical and 3D-spatial relationships with varied attribute bindings.
Compared with prior works, our benchmark is remarkably more challenging, incorporating generation tasks with 3+ generation subjects with complex 3D-spatial relationships.
Additionally, we propose \textbf{CompQuest}, an interpretable and accurate evaluation framework that first decomposes complex prompts into atomic sub-questions, then utilizes a Multimodal Large Language Model to provide fine-grained binary feedback on the correctness of each aspect of generation elements in model-generated images.
This enables precise quantification of alignment between generated images and compositional prompts. 
Furthermore, we propose \textbf{an alignment framework} that uses CompQuest's feedback as preference signals to improve diffusion models' compositional image generation abilities.
Using adjustable per-image preferences, our method is easily scalable and flexible for different tasks.
Using CompAlign, we evaluate 9 state-of-the-art text-to-image models.
Results show that: (1) models remarkable struggle more with compositional tasks with more complex 3D-spatial configurations, and (2) a noticeable performance gap exists between open-source accessible models and closed-source commercial models.
Further empirical study on using CompAlign for model alignment yield promising results: post-alignment diffusion models achieve remarkable improvements in compositional accuracy, especially on complex generation tasks, outperforming previous approach across different compositional settings.
\end{abstract}

\section{Introduction}
\begin{figure}[t]
\vspace{-2em}
\hspace*{-0.5cm}  
\includegraphics[width=1.05\textwidth]{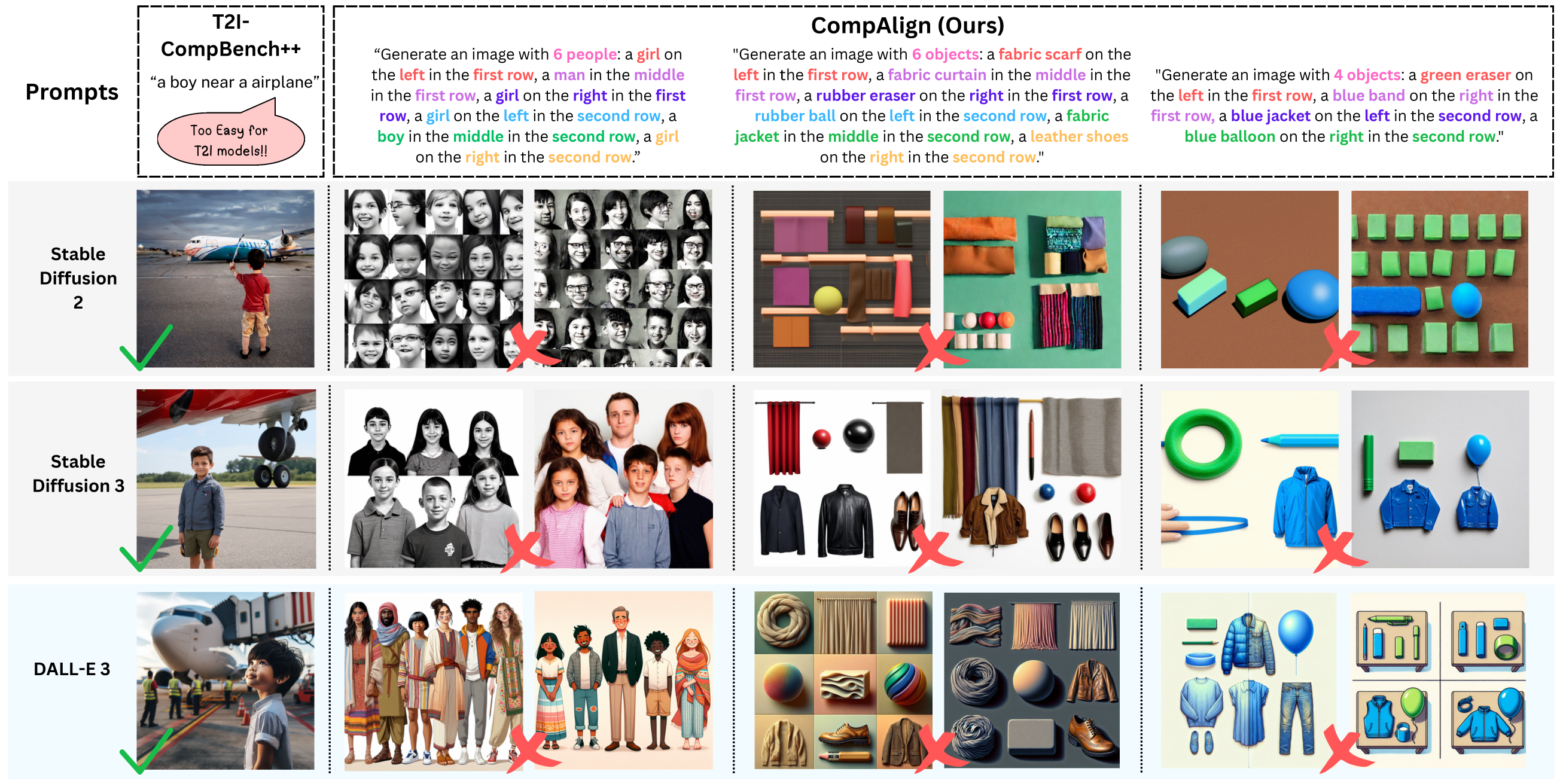}
    \caption{\label{fig:examples_1} Example showing that while T2I models are capable of depicting simple compositional settings in a previous benchmark, they fall short in correctly generating images that accurately conform to complex compositional instructions in our challenging CompAlign benchmark.}
\vspace{-1em}
\end{figure}

Despite impressive recent progress, Text-to-Image (T2I) models still frequently fall short in accurately visualizing details in compositional prompts.
For instance, examples in Figure \ref{fig:examples_1} show how state-of-the-art models like Stable Diffusion 3~\citep{esser2024scaling} and DALL-E 3~\citep{OpenAI_2023} struggle with depicting multiple subjects in compositional instructions involving numeracy, 3D-spatial positionings, and attribute bindings like color and texture.
Such drawbacks in T2I models greatly hinder downstream applications such as design and entertainment, which require better textual control over nuanced and complicated details in generated images.

Prior works~\citep{huang2023t2i, 10847875} proposed approaches to assess compositional T2I abilities, but face 2 major limitations: 
(1) they mostly evaluate models on simple T2I tasks depicting 2 objects, and does not include more challenging settings like depicting 3+ objects with 3D-spatial relationships.
(2) their metrics rely on opaque model-generated scores, lacking both interpretability and robustness in evaluation outcomes.

To bridge these gaps, we introduce \textbf{CompAlign}, a more challenging benchmark for advanced evaluation of compositional T2I generation.
CompAlign consists of 900 complex generation prompts for multi-subject image generation, combining numerical, 3D-spatial, and attribute bindings to reflect realistic complexity in image generation tasks. 
We further propose \textbf{CompQuest}, a vision-language-based evaluation framework that utilizes 3 key steps---\textit{Atomic Question Decomposition}, \textit{Atomic Multimodal Large Language Model (MLLM) Feedback}, and \textit{Aggregated Compositional Accuracy}---to achieve fine-grained evaluation of the compositional image generation task.


Using CompAlign, we conducted evaluation experiments on 9 state-of-the-art T2I models.
Results show that: (1) most models still struggle with accurately depicting compositional scenes, especially for tasks involving multiple generation subjects and complex 3D-spatial positioning configurations; (2) remarkable performance gaps can be observed between open-source and closed-source models.
This indicates that limitations on computational resources and the quality of openly accessible data remain a challenge for building openly accessible generative intelligence.

We further explore methods to improve open-source T2I models on complex compositional generation tasks.
Previous works have leveraged human preference optimization to boost model performance.
However, for compositional images with multiple objects associated with different attributes and relations, relying on simple human preferences is not enough to provide sufficient guidance for diffusion models.
In cases where multiple objects and relationships are present, it might be subjective to decide whether a partially-correct generated image is ``preferred'' over another one.
To address this limitation, we construct \textbf{an alignment framework} that adjustably adopts aggregated atomic feedback from CompQuest as per-image preference signals for improving diffusion models on compositional image generation.
Experimental results with our alignment method on open-weight diffusion models demonstrate substantial improvements in compositional accuracy, outperforming previous methods for improving the compositional abilities of diffusion models.

\section{CompAlign Benchmark}
Assessing the compositionality of T2I models, or the ability to accurately and coherently depict multiple concepts in a unified scene based on textual instructions, has been an increasingly challenging research topic~\citep{10847875}.
For instance, \citet{huang2023t2i,10847875} constructed textual prompts with simple attribute bindings and spatial relationships.
Their evaluation benchmark mostly consists of straightforward generation prompts involving 2 subjects---for example, \textit{``a boy near a airplane''}, as shown in Figure \ref{fig:examples_1}.
However, the figure shows that recent T2I models like SD2 and DALL-E 3 already achieve satisfactory performance on such simple generation tasks.

To further challenge recent stronger T2I models on more complex compositional image generation tasks, we propose \textbf{the CompAlign benchmark}, which consists of a variety of detailed image generation prompts involving multiple objects, 3D-spatial relationships, and precise attributes such as color and texture. 
Our benchmark notably extends the complexity and scope of ~\citet{10847875}'s work through an emphasis on generation tasks for compositional scenarios that depict up to 6 subjects, combined with specific attribute bindings and 3D-spatial positioning requirements.
Such complicated tasks challenge T2I models to understand and accurately depict not only multiple entities with different features, but also the nuanced 3-D positioning of them in the generated scenes. \footnote{We will release our code and data under the CC By 4.0 License upon acceptance.}

\subsection{Prompt Construction}
We adopt a hierarchical heuristic-based pipeline to generate reasonable and diverse compositional image generation prompts.
Specifically, the prompt construction pipeline begins with assigning top-level numerical and 3D-spatial relationships between potential entities.
Then, we iterate through a variety of candidate entities and corresponding attributes to be combined with every assignable 3D-spatial positioning in each generation setting.

\subsubsection{Numerical and 3D-Spatial Relationships}
To create generation settings with complex 3D-spatial positioning of entities, we manually define \textbf{5} distinct 3D-spatial configurations, escalating in both numerical and spatial complexity: 1 row $\times$ 2 subjects, 1 row $\times$ 3 subjects, 2 rows $\times$ 1 subject, 2 rows $\times$ 2 subjects, and 2 rows $\times$ 3 subjects.

We phrase these positional configurations into natural prompt templates for further combination with generation subjects.
For instance, for the configuration with 1 row and 2 subjects, the corresponding prompt template is:

\begin{adjustbox}{minipage=.7\textwidth,margin=0pt \smallskipamount,center}
\vspace{-0.5em}
\small
    \textit{``An image with 2 objects: a \textcolor{cyan}{\{(optional) attribute A\}} \textcolor{violet}{\{subject A\}} on the left, and a \textcolor{cyan}{\{(optional) attribute B\}} \textcolor{violet}{\{subject B\}} on the right.''}
\vspace{-0.5em}
\end{adjustbox}

Specifications of the spatial attributes are detailed in the first section of Appendix \ref{sec:appendix-prompt}, Table \ref{tab:compalign-attributes}. 
The final CompAlign benchmark is equally divided between the 5 numerical and 3D-spatial positioning assignments.
Our established spatial relationship framework can be easily combined with different generation entities and attributes, allowing for easy scaling up of evaluation experiments for compositional T2I generation.

\subsubsection{Entities and Attribute Binding}
Next, we enrich the defined 3D-spatial configuration frameworks with generation subjects and corresponding attribute bindings.

\paragraph{Human Figure Generation}
We begin by integrating the human figure generation task, which we denote as the \textit{``people$\_$only''} category.
Since our 3D-spatial definitions might result in blocked attributes for depicted people, we only combine the human figure generation task with visual gender traits as attributes.

\paragraph{Object Generation}
For object generation tasks, we first incorporate the vanilla \textit{``object$\_$only''} generation settings, in which we directly combine candidate object entities with the defined 3D-spatial positionings.
Then, we follow previous work~\citep{10847875} to consider two major attribute bindings to be further combined with objects: color (\textit{``object\_color''}) and texture (\textit{``object\_texture''}).
Note that we did not include the ``shape'' attribute since such attributes are abstract for automatic evaluation and oftentimes result in unnatural prompts (e.g. ``a rectangular watch'' and ``a teardrop cherry'', as used in~\citet{10847875}'s work.)
To ensure the naturalness of attribute bindings in textual prompts, we utilize the object-attribute pairs defined in~\citet{10847875}'s work.
We provide the holistic lists of all included color and texture attributes in the second section of Table \ref{tab:compalign-attributes}.

\subsubsection{Context-specific object generation} 
To further create challenging yet natural settings for evaluating compositional T2I generation, we additionally integrate explicit scenes for a proportion of evaluation data.
Particularly, we construct object generation settings in kitchens and bathrooms to add further real-world complexity.
For each context, we follow previous work~\citep{10847875} to selectively include objects that naturally appear in the real-life setting (e.g. bowls and cabinets for kitchens), as well as their corresponding natural attribute bindings.
We rephrase our previous generation prompt template for each context-specific object generation task.
For instance, in the kitchen context, the evaluation prompt will be of the format:

\begin{adjustbox}{minipage=.8\textwidth,margin=0pt \smallskipamount,center}
\vspace{-0.5em}
\small
    \textit{``An image with \textcolor{magenta}{\{NUM\}} objects in the kitchen: a \textcolor{cyan}{\{(optional) attribute A\}} \textcolor{violet}{\{subject A\}} on ..., ''}
\vspace{-0.5em}
\end{adjustbox}

\vspace{-0.5em}
\subsection{Benchmark Statistics}
From all possible compositional prompts with combinations of 3D-spatial relationships, generation subjects, attributes, and generation contexts, we sample 900 to construct the final CompAlign benchmark to limit computational costs at training and inference time. 
The 900 entries are specifically curated such that they are evenly distributed among the five 3D-spatial configurations.
Table \ref{tab:compalign-stats} and Figure~\ref{fig:compalign-data} in Appendix \ref{appendix:dataset-statistics} provide a detailed breakdown and visualization of dataset statistics across different categories of generation subjects and 3D-spatial configuration types.
Finally, we adopted a 90-10 train-test split on CompAlign, which is balanced across sub-categories, to better train and evaluate T2I models.
All evaluation experiments are run on the test split at inference time.

\section{Evaluation Framework}
\begin{figure}[t]
\centering
\hspace*{-0.5cm}  
\includegraphics[width=0.92\textwidth]{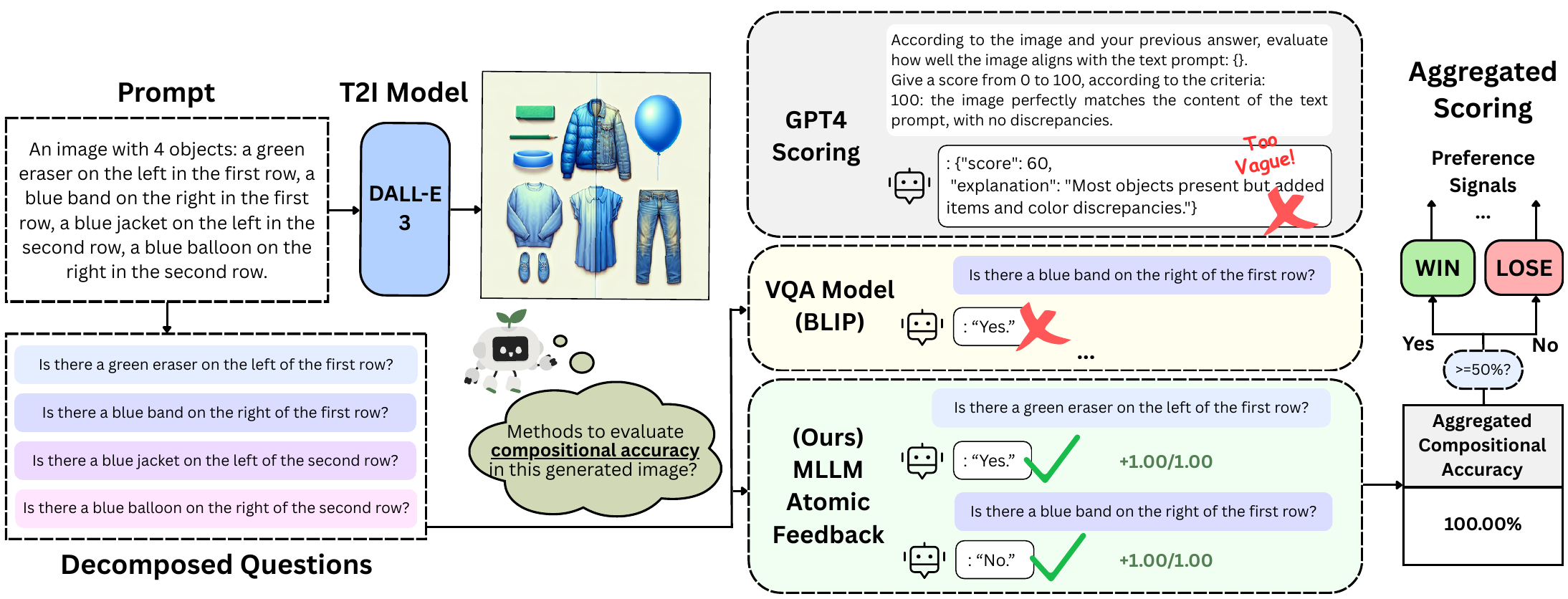}
    \caption{\label{fig:eval_framework} The CompQuest Evaluation Framework. Each compositional prompt is first divided into atomic questions, then answered by a MLLM feedback model, finally aggregated to be an accuracy score.}
\vspace{-1em}
\end{figure}

We propose \textbf{CompQuest}, a Vision-Language-based framework to comprehensively and quantitatively assess compositional T2I generations.
Prior works explored VQA-based metrics~\citep{sun2023dreamsync} or using MLLMs to directly score each generated image~\citep{huang2023t2i, 10847875}.
However, as Figure \ref{fig:eval_framework} demonstrates, both methods fall short in achieving accurate and interpretable evaluation outcomes: VQA models often fail to correctly capture the correctness of compositional attributes in complicated images (e.g. with multiple entities), and directly using MLLM's scoring results in interpretability and robustness issues in the evaluation results.
CompQuest addresses these limitations by leverageing 3 crucial steps during evaluation:

\begin{enumerate}
\vspace{-0.5em}
    \item An \textbf{Atomic Question Decomposition}
 step to divide complex compositional prompts into individual questions that can be judged with binary answers (i.e. ``yes'' or ``no'').
 \vspace{-0.2em}
    \item An \textbf{Atomic MLLM Feedback} step that uses a strong MLLM to judge the depiction accuracy of each decomposed aspect in generated images.
    \vspace{-0.2em}
    \item An \textbf{Aggregated Compositional Accuracy} metric that yields an objective and interpretable quantitative evaluation score for each model generation.
\vspace{-1em}
 \end{enumerate}


\subsection{Atomic Question Decomposition}
Each compositional generation prompt in CompAlign benchmark can be naturally decomposed into multiple atomic questions, each validating (1) the presence, (2) the attribute, and (2) the correct 3D-spatial positioning of target generation subjects. 
For example, Figure \ref{fig:eval_framework} shows an example of how a complex compositional prompt involving 4 objects with 3D-spatial relationships can be decomposed into 4 atomic questions, each verifiable with binary feedback.
This step ensures targeted and interpretable evaluation of each compositional element.

\subsection{Atomic MLLM Feedback}
To verify each atomic question yielded in the previous step, we utilize a strong MLLM to provide binary correctness feedback based on the generated image, indicating clearly whether each generation subject has been correctly depicted and positioned. 
The transparency and determinism of this binary feedback step addresses the issue of interpretable and robustness in previous approach~\citep{10847875} that directly employs MLLMs to output scalar scores.

\subsection{Aggregated Compositional Accuracy}
Finally, we construct the \textbf{Aggregated Compositional Accuracy (ACA)} metric that takes into consideration all atomic-level binary feedback to output a single depiction accuracy score for each image. 
Specifically, ACA is computed as the proportion of correctly-depicted entities, measured by the percentage of atomic questions with positive judgements by the MLLM.
Given a compositional generation prompt $p$ with $n$ target entities $\{e_1, e_2, ..., e_n\}$, we use a T2I model $f_{\theta}$ to generate an image $x = f_{\theta}(p)$ accordingly.
Then, we decompose the prompt into $n$ atomic sub-questions, 1 for verifying each element. 
An MLLM $M$ is adopted to obtain binary feedback for each sub-question:
\begin{equation}
M(q_i, x) = \begin{cases}
1 & \text{if entity } e_i \text{ is correctly depicted in } x, \\
0 & \text{otherwise},
\end{cases}
\end{equation}
where $M(q_i, x)$ represents the positivity of sub-question $q_i$ based on image $x$, as judged by MLLM $M$. 
Then, ACA score for the image can be computed as: \(ACA(x) = \frac{1}{n}\sum_{i=1}^{n} M(q_i, x).\)

ACA provides an intuitive measure of T2I models' capability to tackle complex compositional image generation tasks.
In the next section, we also illustrate how ACA can be further adopted to facilitate the improvement of diffusion-based T2I models.

\section{Improving Diffusion Models with CompAlign}
\label{sec:alignment}
We further explore how we can improve open-weight T2I diffusion models on challenging compositional image generation tasks.
It is intrinsically difficult to obtain ground truth images specifically for each complex compositional prompt as direct fine-tuning data.
Traditional reward training requires pairwise preference data (i.e. 1 preferred and 1 not preferred image for the same prompt) and human preference feedback, which creates additional challenges.
For complicated compositional tasks (e.g. involving multiple objects), especially involving partially-correct images, it is costly and naturally hard for human annotators to output robust and accurate preferences between generations.

In their previous work, ~\citet{lialigning} leverages binary per-image human feedback as preference for aligning diffusion models, enabling easy scalability of alignment data.
We extended their approach to automatically construct per-image binary preferences based on the MLLM feedback on CompAlign data, using our CompQuest evaluation framework.
Then, we utilize the binary preference signals for model-generated images on CompAlign to align diffusion models for better T2I compositionality.
Not only does our approach address the limitations from the reliance on human feedback, but it also provides a flexible alignment framework in which preference criteria can be adjusted subject to different compositional tasks.

\subsection{Automatic Binary Feedback via CompQuest}
We utilize CompQuest's ACA scores to construct per-image binary preference signals. 
Specifically, we define an image to be a ``win'' if its ACA score exceeds a predefined threshold $\tau$:
\begin{equation}
\small
w(x) = \begin{cases}
1 & \text{if } ACA(x) \ge \tau, \\
-1 & \text{otherwise}.
\end{cases}
\end{equation}
The threshold $\tau$ can be adjusted for compositional T2I tasks with different difficulty levels, allowing for a more flexible and nuanced alignment strategy compared to rigid binary feedback from humans or traditional pairwise preference data.

\subsection{Training with Automated Binary Feedback}
Next, we utilize the ACA-based win/lose signals to fine-tune diffusion models via a similar binary classification objective as in~\citet{lialigning}'s work.
Following~\citet{ethayarajh2024kto} and ~\citet{lialigning}, we optimize the sampling policy $\pi_\theta(x_{t-1} \mid x_t)$ based on the whether a generation is considered a ``win'' or a ``lose''.
\begin{equation}
\max_{\pi_\theta} \mathbb{E}_{x_0 \sim \mathcal{D}, \, t \sim \text{Uniform}([0, T])} 
\left[
    U\left(w(x_0)\left(\beta \log \frac{\pi_\theta(x_{t-1} \mid x_t)}{\pi_{\text{ref}}(x_{t-1} \mid x_t)} - Q_{\text{ref}} \right)\right)
\right]
\end{equation}
where $U(\cdot)$ is a strictly-increasing value function mapping implicit reward to subjective utility.
The reference point $Q_{\text{ref}} = \beta \, \mathbb{D}_{\text{KL}} \left[ \pi_\theta(a\mid s) \| \, \pi_{\text{ref}}(a\mid s) \right]$ for an action $a$, state $s$, and divergence penalty control hyperparameter $\beta$.

\vspace{-0.5em}
\subsection{Augmented Alignment Data}
Additionally, to further enhance the effectiveness of alignment on challenging compositional T2I tasks, we explore the augmentation of alignment data generated by stronger T2I models. 
Specifically, we employ compositional images generated from Stable Diffusion (SD) 3.5~\citep{esser2024scaling} and DALL-E 3~\citep{OpenAI_2023} to augment alignment data on CompAlign.
This approach is capable of bootstraping weaker diffusion models, which might originally lack compositional ability, toward better understanding and performance on complex generation tasks. 
The augmentation approach is also easily scalable to more data and T2I models in a model structure-agnostic way.

\begin{table}[t]
    \vspace{-1em}
    \centering
    \renewcommand*{\arraystretch}{1.0}
    \scriptsize
    \begin{tabular}{p{0.14\textwidth}|p{0.03\textwidth}p{0.03\textwidth}p{0.04\textwidth}p{0.055\textwidth}p{0.075\textwidth}p{0.05\textwidth}|p{0.03\textwidth}p{0.03\textwidth}p{0.03\textwidth}p{0.03\textwidth}p{0.035\textwidth}|p{0.04\textwidth}}
    \toprule
    \midrule
    \multirow{6}*{\textbf{Model}} & \multicolumn{12}{c}{\textbf{Average Compositional Accuracy}} \\
    \cmidrule{2-13}
     & \multicolumn{6}{c}{\textbf{Generation Type}} & \multicolumn{5}{c}{\textbf{3D-Spatial Configuration}} &  \\
    \cmidrule{2-13}
      & \textbf{people only} & \textbf{object only} & \textbf{object color} & \textbf{object texture} & \textbf{object color \;\; bathroom} & \textbf{object color kitchen} & \textbf{1row $\times$ 2sub} & \textbf{1row $\times$ 3sub} & \textbf{2rows $\times$ 1sub} & \textbf{2rows $\times$ 2sub} & \textbf{2rows $\times$ 3sub}  & \multirow{3}*{\textbf{Overall}} \\
    \midrule
    \midrule
    \multicolumn{13}{c}{\textit{Open-Source Models}} \\
    \midrule
    SD1.5 & 62.50 & 47.46 & 34.78  & 36.23 & 59.52 & 73.81 &  53.97 & 44.44  &  50.00 &  35.00 & 33.33  & 43.31 \\
    SD2 & 86.67 & 47.67 & 40.33 & 39.00 & 41.90 & 57.14 & 63.19 & 43.06 & 61.11 & 31.67 & 29.37  & 46.28 \\
    SDXL & 93.33 & 62.00 & 41.67 & 53.00 & 76.19 & 73.33  &  72.22 & 54.17 & 72.22 & 43.33 & 44.31 & 57.01\\
    SD3  & 83.33 & 76.33 & 80.33 & 73.33 & 81.90 & 79.05 & 93.06 & 87.50 & 61.11 & 63.33 & 63.23  & 77.46 \\
    SD3.5  & 86.67 & 80.67 & 80.33 & 88.67 & 75.24 & 78.10 & 92.36 & 84.72 & 94.44 & 65.00 & 75.93 & 82.69 \\
    Flux.1 & 86.67 & 88.33 & 83.33 & 81.67 & 75.24 & 81.90 & 87.50 & 86.11 & 72.22 & 90.00 & 76.98  & 83.92 \\
    \midrule
    \multicolumn{13}{c}{\textit{Closed-Source Models}} \\
    \midrule
    DALL-E 3  & 90.00 & 90.00 & 76.00 & 78.00 & 73.33 & 82.86 & 96.53 & 81.94 & 88.89 & 73.33 & 63.76  & 81.46 \\
    gpt-image-1-high & 93.33 & \textbf{94.33} & \textbf{92.00} & \textbf{94.67} & 84.76 & \textbf{100.00} & \textbf{98.61} & \textbf{91.67} & \textbf{94.44} & \textbf{88.33} & 92.99 & \textbf{93.51} \\
    Gemini  & \textbf{100.00} & 91.67 & 89.33 & 88.67 & \textbf{97.14} & 94.29 & 91.67 & 87.95 & 90.28 & 83.33 & \textbf{95.45} & 90.68 \\
    \midrule
    \bottomrule
    \end{tabular}
    \vspace{0.2em}
    \small
    \captionof{table}{\label{tab:strat-results-1} 
    Comprehensive evaluation results across 9 open- and closed-source T2I models. Closed-source commercial models achieve remarkably stronger compositional accuracy, with OpenAI's gpt-image-1-high as the leading model.
    }
    \vspace{-2em}
\end{table}

\vspace{-0.5em}
\section{Experiments}
\vspace{-0.5em}
We conduct 2 major empirical studies in this work.
We first utilize the proposed CompAlign benchmark to assess the compositional ability of T2I models.
Next, we utilize the proposed CompAlign framework to improve the compositional performance of open-source diffusion models.

\subsection{Evaluation}
Utilizing the test set of CompAlign, we evaluate \textbf{9} state-of-the-art T2I models on the compositional image generation tasks.
These include \textbf{6 open-source models}---SD1.5, SD 2~\citep{Rombach_2022_CVPR}, SD XL, SD3~\cite{esser2024scaling}, SD3.5, and Flux.1~\citep{flux2024}---as well as \textbf{3 closed-source models}, including OpenAI's DALL-E 3~\citep{OpenAI_2023}, gpt-image-1-high, and Google's Gemini~\citep{team2023gemini}.

\paragraph{Implementation Details}
We follow the default implementations of the open-source models, using the Huggingface Diffusers' \textit{DiffusionPipeline} for SDXL, \textit{StableDiffusionPipeline} for SD2, SD3, and SD3.5, and \textit{FluxPipeline} for Flux.1.
For both closed-source models, we utilize the API services provided by OpenAI and Google with default configurations.
For the evaluation MLLM, we use the \textit{``gpt-4o-mini-2024-07-18''} model to provide binary feedback for each aspect of compositional generation.
Details and prompts are provided in Appendix \ref{appendix:full_results}.

\begin{table}[h]
    \vspace{-1.0em}
    \centering
    \renewcommand*{\arraystretch}{1.0}
    \scriptsize
    \begin{tabular}{p{0.14\textwidth}|p{0.03\textwidth}p{0.03\textwidth}p{0.04\textwidth}p{0.055\textwidth}p{0.075\textwidth}p{0.05\textwidth}|p{0.03\textwidth}p{0.03\textwidth}p{0.03\textwidth}p{0.03\textwidth}p{0.035\textwidth}|p{0.04\textwidth}}
    \toprule
    \midrule
    \multirow{6}*{\textbf{Model}} & \multicolumn{12}{c}{\textbf{Average Compositional Accuracy}} \\
    \cmidrule{2-13}
     & \multicolumn{6}{c}{\textbf{Generation Type}} & \multicolumn{5}{c}{\textbf{3D-Spatial Configuration}} &  \\
    \cmidrule{2-13}
      & \textbf{people only} & \textbf{object only} & \textbf{object color} & \textbf{object texture} & \textbf{object color \;\; bathroom} & \textbf{object color kitchen} & \textbf{1row $\times$ 2sub} & \textbf{1row $\times$ 3sub} & \textbf{2rows $\times$ 1sub} & \textbf{2rows $\times$ 2sub} & \textbf{2rows $\times$ 3sub}  & \multirow{3}*{\textbf{Overall}} \\
    \midrule
    \midrule
    SD1.5 & 62.50 & 47.46 & 34.78  & 36.23 & 59.52 & 73.81 & 53.97 & 44.44 & 50.00 & 35.00 & 33.33 & 43.31 \\
    \;\; + CompAlign & 66.67 & 49.67 & 31.00 & 45.67 & 70.48 & 76.19 & 63.89 & 43.06 & 61.11  & 31.67  & 35.19 & 46.94 \\
    \midrule
    SD2 & 86.67 & 47.67 & 40.33 & 39.00 & 41.90 & 57.14 & 63.19 & 43.06 & 61.11 & 31.67 & 29.37 & 46.28 \\
    \;\; + CompBench++ & 83.33 & 44.33 & 28.67 & 41.33 & 57.14 & \textbf{62.86} & 54.17 & 40.28 & 55.56 & 25.00 & 40.74 & 45.41 \\
    \;\; + CompAlign & \textbf{90.00} & \textbf{56.00} & \textbf{42.33} & \textbf{51.33} & \textbf{57.14} & 60.00 & \textbf{68.75} & \textbf{54.17} & \textbf{61.11} & \textbf{35.00} & \textbf{41.80} & \textbf{53.08} \\
    \midrule
    \bottomrule
    \end{tabular}
    \vspace{0.2em}
    \captionof{table}{\label{tab:strat-results-2}
    Experiment results on CompAlign-ed diffusion models, compared with the base models and the best-performing checkpoint in previous works. CompAlign effectively and consistently improves the performance for both base diffusion models on compositional tasks, especially in complex generation settings.
    }
\vspace{-1.2em}
\end{table}

\paragraph{Evaluation Results}
Table \ref{tab:strat-results-1} reports investigated models' performances on compositional T2I generation, as measured in Average Compositional Accuracy (ACA).
We stratify evaluation outcomes in terms of: (1) generation subject types, and (2) 3D-spatial configuration types.
Additionally, we report the overall ACA performance for each model in the last column.

\begin{figure}[ht]
\vspace{-0.5em}
\centering
\hspace*{-1cm}  
\includegraphics[width=1.1\textwidth]{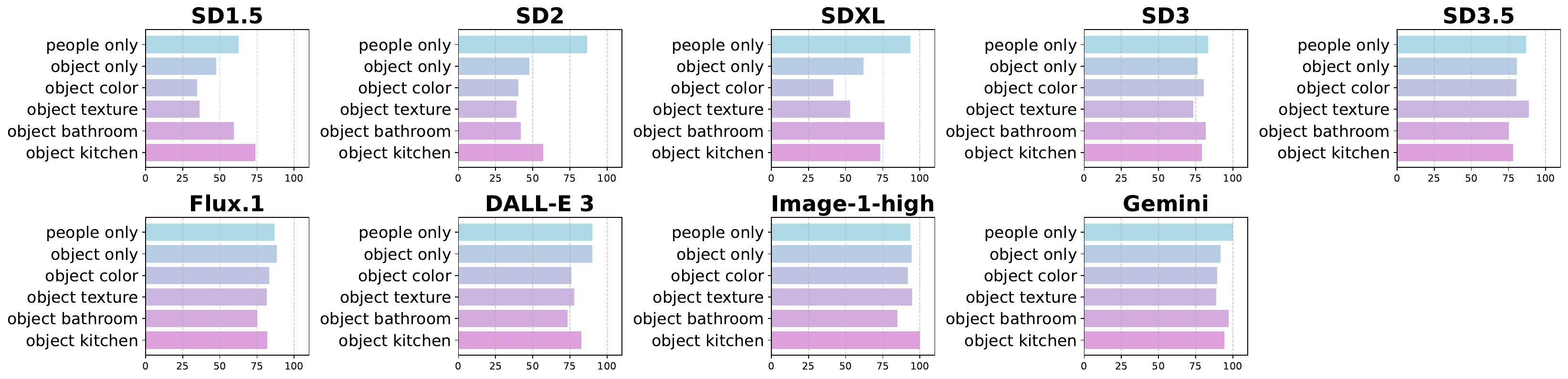}
    \caption{\label{fig:strat-aca} Visualization of evaluation results, stratified by generation types. We observe that the older generation of UNet-based diffusion models fall short in achieving good compositional performance for tasks such as depicting the color attribute bindings for objects.}
\vspace{-0.5em}
\end{figure}

As shown in Table \ref{tab:strat-results-1} and Figure \ref{fig:strat-aca}, open-source diffusion models with U-Net structures, such as SD1.5 and SD2, exhibit limited compositional accuracy (below 50\%). 
Performance improves with model scale, generation, and architecture improvements.
For instance, SD3 and SD3.5, which utilize a transformer-based structure to replace U-Nets, both achieve remarkably higher ACA scores than the previous two, reaching above 75\%. 
Among closed-source models, OpenAI's gpt-image-1-high model leads with 93.51\%, outperforming DALL-E 3 and Gemini. 
While results show promising advances in architectural and training pipelines for T2I models, we also observe a \textbf{remarkable performance gap between open-sourced and closed-source models}.

Stratified results by 3D-Spatial Configuration Types also reveal interesting trends.
We observe that \textbf{model performance generally degrades with increasing spatial complexity}. 
For instance, while DALL-E 3 achieves an accuracy of as high as 96.53\% when depicting the simple ``1 row $\times$ 2 subjects'' spatial relationship, its ACA drops to 63.76 for the complicated ``2 rows $\times$ 3 subjects'' task.
The older generation of open-source diffusion models like SD1.5 and SD2 struggle across all configurations, especially on two-row settings. 
In contrast, larger and more recent models like SD3.5 and Flux.1 show significant improvements on complex spatial compositionality. 
Closed-source models remain dominant: DALL-E 3 and gpt-image-1-high achieve near-perfect scores on simpler layouts and maintain high accuracy in complex settings. 
This further highlights \textbf{the performance gap between closed-source and open-source models on compositional T2I generation}.

\subsection{CompAlign-ing T2I Diffusion Models}
Using the CompQuest-based alignment framework specified in Section \ref{sec:alignment}, we conducted experiments to improve diffusion models' compositional T2I abilities.

\vspace{-0.5em}
\paragraph{Implementation Details}
Building upon~\citet{lialigning}'s image-wise preference optimization framework, we aligned 2 open-source diffusion models with U-Net structures: SD 1.5 and SD 2.
For alignment data augmentation, we employed DALLE-3 and SD3.5's generations on the training set of CompAlign to construct the per-image binary preference data; a generated image is counted as a ``win'' if its ACA surpasses 0.5, and it is counted as a ``lose'' otherwise.
At inference time, we report ACA outcomes on the held-out test set.
We train both models on 2 NVIDIA A6000 GPUs with a batch size of 4 each GPU using Adam optimizer. 
We set the KL Divergence penalty term $\beta$ to 1000 and the probability $\gamma$ to sample from desirable samples to 0.8.
We adopt a base learning rate of 1e-7 to train for 10000 iterations.

\begin{figure}[t]
\vspace{-1.5em}
\hspace*{-0.5cm}  
\includegraphics[width=1.05\textwidth]{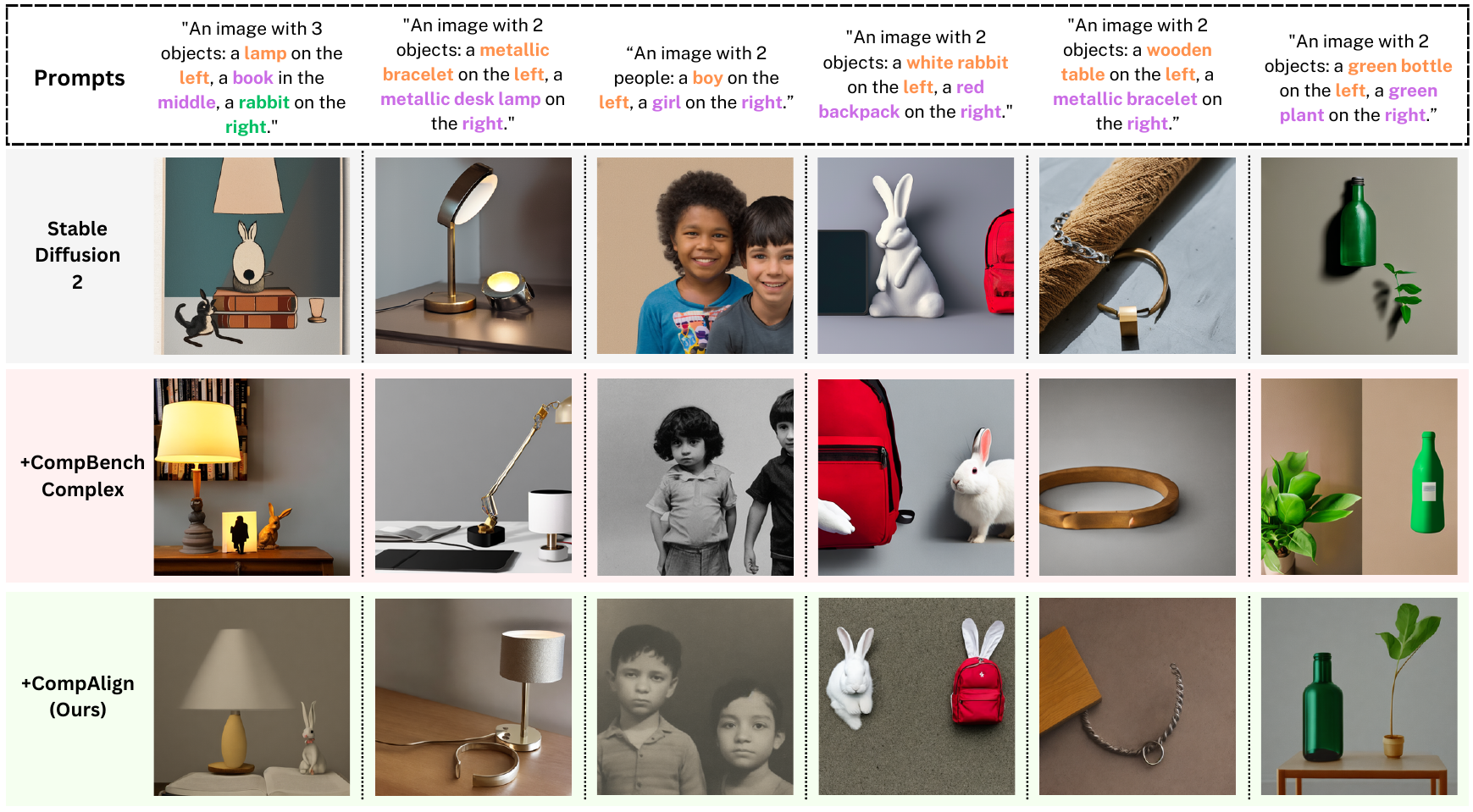}
    \caption{\label{fig:examples_2} Qualitative examples of how CompAlign-ed diffusion model demonstrate better performance on compositional T2I tasks.}
\vspace{-1.5em}
\end{figure}

\begin{figure}[b]
\vspace{-1.8em}
\centering
\begin{minipage}[t]{.45\linewidth}
    \vspace{0.5em}
    \hspace*{-1cm}
    \centering
    \includegraphics[width=0.9\textwidth]{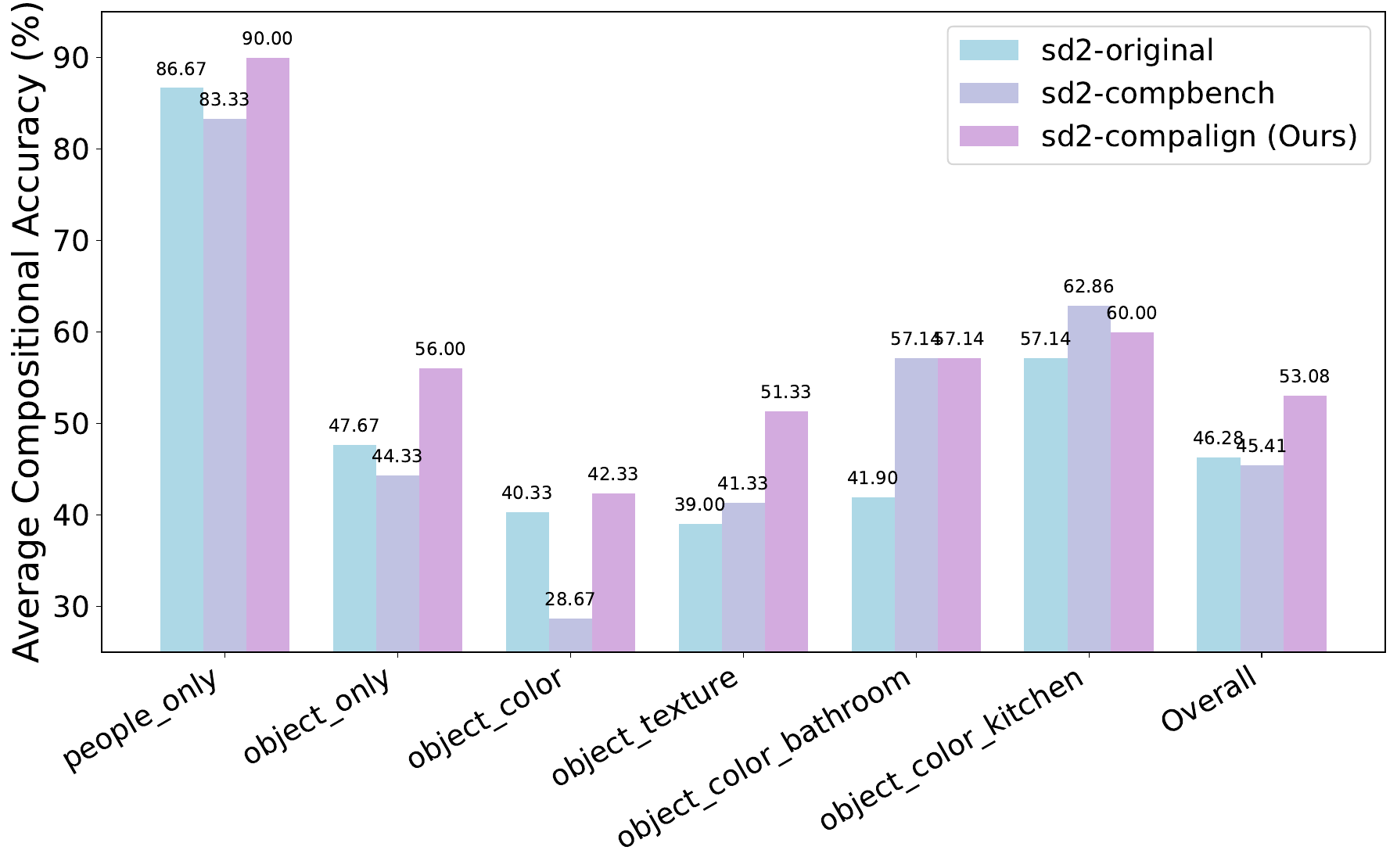}
    \vspace{-0.5em}
\end{minipage} 
\hspace*{0.2em}
\begin{minipage}[t]{.45\linewidth}
    \vspace{0.5em}
    \hspace*{-0.8cm}
    \centering
    \includegraphics[width=0.9\textwidth]{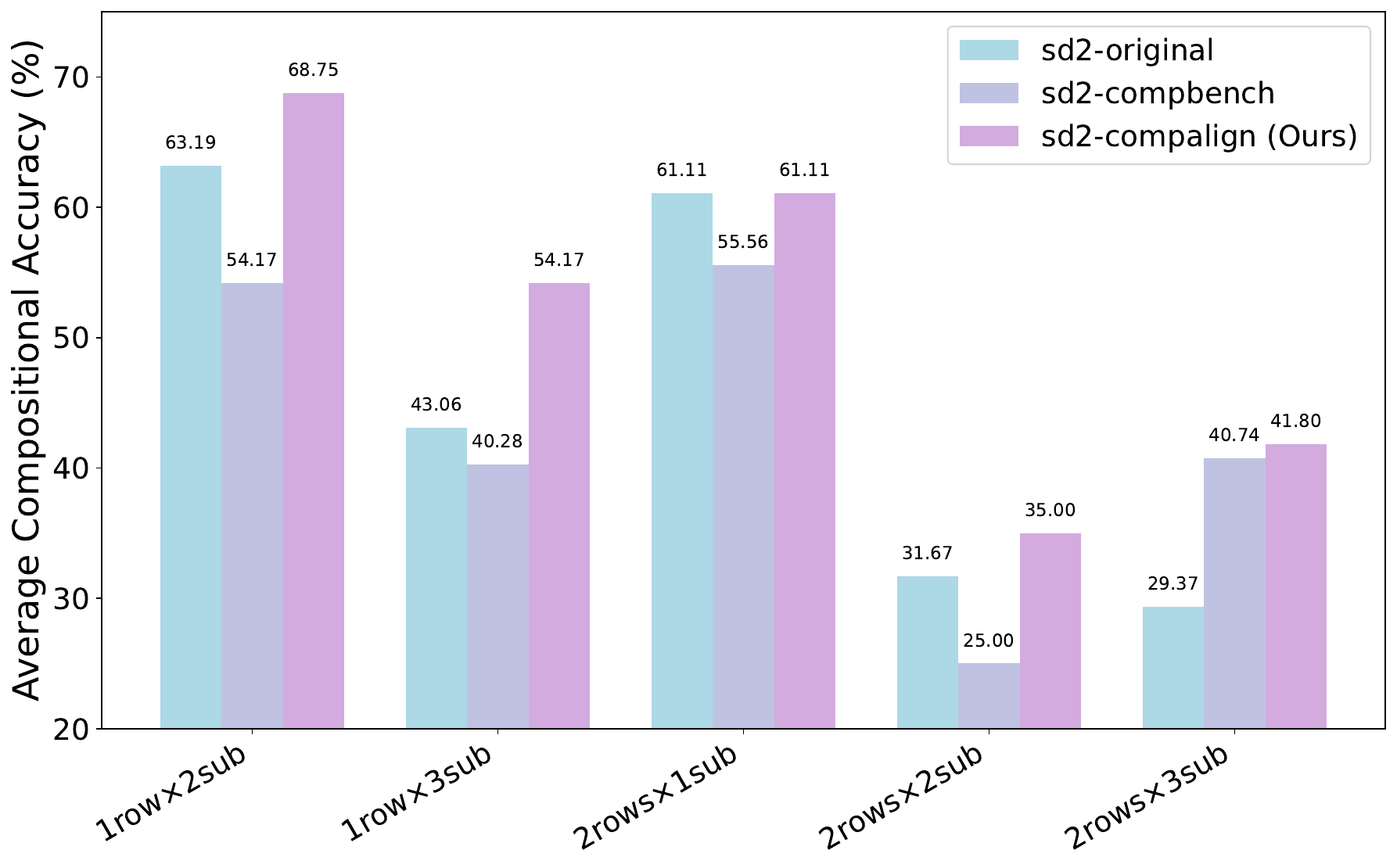}
    \vspace{-0.5em}
\end{minipage}
\caption{\label{fig:align-res-2} Visualization of experiment results on SD2. Our alignment method effectively improves the compositional ability of diffusion models on most generation aspects, whereas ~\citet{10847875}'s checkpoint frequently underperforms even the base model.}
\vspace{-2.0em}
\end{figure}

\vspace{-0.5em}
\paragraph{Baselines}
~\citet{10847875} released several model checkpoints using their proposed training method to improve compositional T2I ability of the SD2 model.
We compare out method with all of their provided checkpoints, with full results reported in Appendix \ref{appendix:full_results}.
In our main evaluation results, we only list out the outcomes of their best overall-performing checkpoint as measured by ACA.

\vspace{-0.5em}
\paragraph{Evaluation Results}
As shown in Table ~\ref{tab:strat-results-2}, CompAlign consistently improves compositional accuracy of both SD1.5 and SD2 across generation types and 3D-spatial configuration types. 
Figure \ref{fig:align-res-2} visualizes stratified results across different generation settings for more straightforward comparison between models.
For SD1.5, applying CompAlign yields notable improvements in the compositional object generation task with texture attribute bindings.
For both models, CompAlign boosts performance on compositional tasks involving complex 3D-spatial configurations, such as 2rows×1subject (+11.11\%). 
Notably, CompBench++ fails to produce consistent performance gains on the compositional task, even under-performing the original SD2 model on generation tasks like object with color attribute bindings. 
Empirical results across different evaluation dimensions and base models demonstrate the effectiveness of our compositionality-aware alignment strategy.
Additionally, Figure \ref{fig:examples_2} provides qualitative examples of accurate compositional images generated by CompAlign-ed SD2 model, versus the incorrect generations of the base SD2 model and~\citet{10847875}'s checkpoint.

\vspace{-0.5em}
\section{Related Works}
\vspace{-0.5em}
\subsection{Compositional Text-to-Image Generation}
\vspace{-0.5em}
Previous studies have made valuable efforts in exploring the limitations of current T2I models in understanding and depicting compositionality in generation tasks.
For instance, \citet{feng2022training}, \citet{chefer2023attend}, and \citet{NEURIPS_DATASETS_AND_BENCHMARKS2021_0a09c884} explored the depiction of objects with color attributes; \citet{lian2023llm} studied the problem of numeracy in generated images; \citet{wu2023harnessing} and \citet{chen2024training} investigated simple spatial relationship (e.g. ``on top of'') in compositional T2I generation.
However, ~\citet{10847875} pointed out that prior works are limited to studying single aspects of sub-problems and constrained evaluation settings.

The most related to our work, ~\citet{huang2023t2i} and ~\citet{10847875} proposed T2I-CompBench++, a benchmark introducing 4 categories and 8 sub-categories of compositionality in their evaluation prompts, such as attribute binding and simple spatial relationships.
However, their benchmark mostly assesses T2I models on generation tasks with only up to 2 entities, and the spatial relationships that they cover still lack 3D-complexity.

\vspace{-0.5em}
\subsection{Aligning Diffusion Models}
To better align T2I diffusion models' generations with preference objectives like human feedback, previous researchers have adopted approaches such as reward model-based fine-tuning to improve diffusion models~\citep{fan2023optimizing,hao2023optimizing,blacktraining,xu2023imagereward,clarkdirectly,prabhudesai2023aligning,lee2023aligning,fan2023dpok,sun2023dreamsync}. 
Specifically, \citet{xu2023imagereward, clarkdirectly,prabhudesai2023aligning}, and~\citet{lialigning} explored methods to align diffusion models with human preferences.
Among these works, \citet{lialigning} introduced a novel alignment framework that overcomes previous works' limitations of relying on pairwise preference data.
They propose to adopt per-image binary feedback as preference signals, broadening the application of alignment in diffusion model improvement.
However, prior works mostly focused on improving general text-image fidelity, and did not extend to more precise control of compositionality (e.g. both numerical and 3D-spatial positioning) in generated images.
Additionally, previous studies mostly relied on human feedback for generated images, which is resource-costly and unreliable in complicated generation settings.

\citet{hu2023tifa} explored the usage of a Visual Question Answering (VQA) model to provide automated faithfulness evaluation for generated images.
Building upon their work, \citet{sun2023dreamsync} was among the first to explore the usage of VQA-based feedback as preference signals for aligning diffusion models. 
However, their work is also limited to the general faithfulness of T2I models, and does not investigate compositionality generation tasks.
Additionally, VQA models fall short in correctly providing feedback for complex compositional generation tasks, as shown in Figure~\ref{fig:eval_framework}.

\section{Conclusion}
In this paper, we propose \textbf{CompAlign}, a challenging evaluation benchmark consisting of $900$ complex textual prompts combining numeracy, 3D-spatial relationships, and attribute bindings, to assess and improve compositional image generation abilities of T2I models.
Along with our benchmark, we introduce the \textbf{CompQuest} evaluation framework, which adopts 3 key steps to first decompose complex generation settings into atomic questions, then obtaining precise atomic feedback from an MLLM, and finally outputting an aggregated compositional accuracy score.
Compared to previous evaluation frameworks, CompQuest is more interpretable and robust in providing accurate feedback on compositional tasks.
Evaluation outcomes on 9 T2I models unveil noticeable performance limitations of open-sourced models.
We further explore and propose the use of CompQuest's feedback as preference signals to align open-source diffusion models for better compositional T2I performance.
Empirical results show that our method is able to consistently and effectively improve the compositional performance of 2 base diffusion models.
Furthermore, our method leverages an automated pipeline to yield per-image binary signals as preference, and therefore is easily scalable and generalizable to other data and T2I tasks.

\bibliographystyle{abbrvnat}
\bibliography{ref}


\appendix
\section{Details on Prompt Construction for CompAlign}
\label{sec:appendix-prompt}
We hereby provide additional details on prompt construction for the CompAlign benchmark.
Table \ref{tab:compalign-attributes} lists all specifications for attributes leveraged to construct prompts.
Note that not all combinations between attributes would be natural---for instance, ``a blue cherry'' wouldn't make sense.
Therefore, we follow the attribute bindings for different objects in ~\citet{10847875}'s work and adopt them when constructing CompAlign data.

\begin{table}[h]
    \centering
    \renewcommand*{\arraystretch}{1.0}
    \scriptsize
    \begin{tabular}{p{0.15\textwidth}p{0.13\textwidth}p{0.65\textwidth}}
    \toprule
    \midrule
    \textbf{Aspect} & \textbf{Category} & \textbf{Attributes} \\
    \midrule
     \multirow{7}*{\textbf{3D-Spatial Relations}} & 1 row $\times$ 2 subjects  & 'on the left', 'on the right' \\
     & 1 row $\times$ 3 subjects  & 'on the left', 'in the middle', 'on the right' \\
     & 2 row $\times$ 1 subject  & 'in the front', 'in the back' \\
     & 2 row $\times$ 2 subjects  & 'on the left in the first row', 'on the right in the first row', 'on the left in the second row', 'on the right in the second row' \\
     & 2 row $\times$ 3 subjects  & 'on the left in the first row', 'in the middle in the first row', 'on the right in the first row', 'on the left in the second row', 'in the middle in the second row', 'on the right in the second row' \\
     \midrule
    \multirow{3}*{\textbf{Generation Subjects}} & people  & 'male', 'female' \\
     & color  & 'red', 'orange', 'yellow', 'green', 'blue', 'purple', 'black', 'white', 'brown', 'pink', 'gray', 'gold', 'silver' \\
     & texture  & 'rubber', 'plastic', 'metallic', 'wooden', 'fabric', 'fluffy', 'leather', 'glass' \\
    \midrule
    \bottomrule
    \end{tabular}
    \vspace{0.2em}
    \captionof{table}{\label{tab:compalign-attributes}
    Attributes included in the CompAlign benchmark.
    }
    \vspace{-0.5em}
\end{table}

\section{Dataset Statistics}
\label{appendix:dataset-statistics}
\begin{figure}[tb]
\vspace{-1em}
\begin{minipage}[t]{.42\linewidth}
    \vspace{0.5em}
    \centering
    \includegraphics[width=\textwidth]{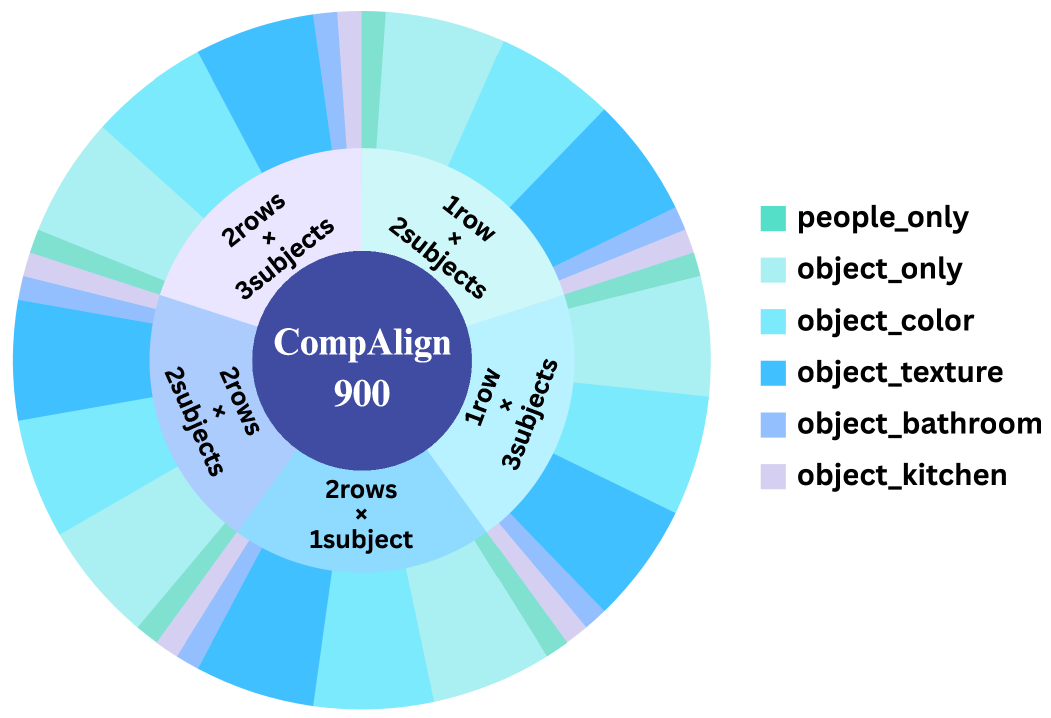}
    \captionof{figure}{\label{fig:compalign-data} CompAlign's benchmark composition.}
    \vspace{-1em}
\end{minipage} 
\hspace*{2em}
\begin{minipage}[t]{.48\linewidth}
    \vspace{0.5em}
    \centering
    \renewcommand*{\arraystretch}{1.0}
    \scriptsize
    \begin{tabular}{p{0.35\textwidth}p{0.35\textwidth}p{0.15\textwidth}}
    \toprule
    \midrule
    \textbf{Aspect} & \textbf{Category} & \textbf{\# Entries} \\
    \midrule
    \textbf{Overall}  & - & 900\\
    \midrule
     \multirow{5}*{\textbf{3D-Spatial Relations}} & 1 row $\times$ 2 subjects  & 180 \\
     & 1 row $\times$ 3 subjects  & 180 \\
     & 2 row $\times$ 1 subject  & 180 \\
     & 2 row $\times$ 2 subjects  & 180 \\
     & 2 row $\times$ 3 subjects  & 180 \\
     \midrule
    \multirow{6}*{\textbf{Generation Subjects}} & people\_only  & 50 \\
     & object\_only  & 250 \\
     & object\_color  & 250 \\
     & object\_texture  & 250 \\
     & object\_bathroom  & 50 \\
     & object\_kitchen  & 50 \\
    \midrule
    \bottomrule
    \end{tabular}
    \vspace{0.1em}
    \captionof{table}{\label{tab:compalign-stats}
    Statistics of the CompAlign benchmark.
    }
\end{minipage}
\vspace{-0.5em}
\end{figure}

We hereby provide additional statistics for the CompAlign data.
Table \ref{tab:compalign-stats} lists out the data composition of each sub-category of CompAlign.
Figure \ref{fig:compalign-data} further visualizes the subdivisions of data.

\section{Full Experiment Details}
\label{appendix:full_results}
\subsection{Details on MLLM for Compositional Accuracy Evaluation.}
After decomposing each compositional T2I prompt into a series of atomic questions judgeable by a binary answer, we adopt the \textit{gpt-4o-mini-2024-07-18} model to return answers to a batch of questions for a generated image.
Evaluation prompt used is listed in Table \ref{tab:prompt-aca}.
We then conduct post-processing on the output of this MLLM judge to obtain the aggregated compositionally accuracy score.

\begin{table}[h]
    \centering
    \begin{tabular}{p{0.8\textwidth}}
    \toprule
    \midrule
    \textbf{Compositional Correctness Evaluation Prompt} \\
    \midrule
    "Given an image and a list of questions, generate a separate yes or no response for each question according to the image. Example response: \\
    \;\;\;\;\;\;\;\; \{ \\
    \;\;\;\;\;\;\;\;\;\;\;\;\;\;\;\;      "responses": \{ \\
    \;\;\;\;\;\;\;\;\;\;\;\;\;\;\;\; \;\;\;\;\;\;\;\;          question 1: 'Yes', \\
    \;\;\;\;\;\;\;\;\;\;\;\;\;\;\;\; \;\;\;\;\;\;\;\;          question 2: 'No' \\
    \;\;\;\;\;\;\;\;\;\;\;\;\;\;\;\;      \} \\
    \;\;\;\;\;\;\;\; \}.  \\
    Questions: " \\
     \bottomrule
    \end{tabular}
    \vspace{1em}
    \caption{Prompt for compositional accuracy evaluation.}
    \label{tab:prompt-aca}
\end{table}

\subsection{Full Experiment Results}
Table \ref{tab:strat-results-full} presents full evaluation results of our alignment framework, base diffusion models, and all checkpoints trained for different compositional T2I abilities, which are provided in ~\citet{10847875}'s work.
In the main paragraphs, we report results for ~\citet{10847875}'s checkpoint on color attribute compositionally, since this checkpoint achieves the best ACA score compared to other checkpoints in the same study.
However, we still observe that the checkpoints in this previous work perform poorly, even frequently underperforming the base SD2 model.

\begin{table}[h]
    \vspace{-0.5em}
    \centering
    \renewcommand*{\arraystretch}{1.0}
    \scriptsize
    \begin{tabular}{p{0.16\textwidth}|p{0.03\textwidth}p{0.03\textwidth}p{0.04\textwidth}p{0.055\textwidth}p{0.075\textwidth}p{0.05\textwidth}|p{0.03\textwidth}p{0.03\textwidth}p{0.03\textwidth}p{0.03\textwidth}p{0.035\textwidth}|p{0.04\textwidth}}
    \toprule
    \midrule
    \multirow{6}*{\textbf{Model}} & \multicolumn{12}{c}{\textbf{Average Compositional Accuracy}} \\
    \cmidrule{2-13}
     & \multicolumn{6}{c}{\textbf{Generation Type}} & \multicolumn{5}{c}{\textbf{3D-Spatial Configuration}} &  \\
    \cmidrule{2-13}
      & \textbf{people only} & \textbf{object only} & \textbf{object color} & \textbf{object texture} & \textbf{object color \;\; bathroom} & \textbf{object color kitchen} & \textbf{1row $\times$ 2sub} & \textbf{1row $\times$ 3sub} & \textbf{2rows $\times$ 1sub} & \textbf{2rows $\times$ 2sub} & \textbf{2rows $\times$ 3sub}  & \multirow{3}*{\textbf{Overall}} \\
    \midrule
    \midrule
    SD1.5 & 62.50 & 47.46 & 34.78  & 36.23 & 59.52 & 73.81 & 53.97 & 44.44 & 50.00 & 35.00 & 33.33 & 43.31 \\
    \;\; + CompAlign & 66.67 & 49.67 & 31.00 & 45.67 & 70.48 & 76.19 & 63.89 & 43.06 & 61.11  & 31.67  & 35.19 & 46.94 \\
    \midrule
    SD2 & 86.67 & 47.67 & 40.33 & 39.00 & 41.90 & 57.14 & 63.19 & 43.06 & 61.11 & 31.67 & 29.37 & 46.28 \\
    \;\; + CB++(color)     & 83.33 & 44.33 & 28.67 & 41.33 & 57.14 & 62.86 & 54.17 & 40.28 & 55.56 & 25.00 & 40.74 & 45.41 \\
    \;\; + CB++(complex)   & 56.67 & 46.33 & 33.33 & 37.00 & 51.43 & 60.00 & 54.17 & 40.28 & 61.11 & 23.33 & 32.80 & 41.98 \\
    \;\; + CB++(nonspatial) & 66.67 & 41.33 & 39.00 & 42.00 & 51.43 & 60.00 & 54.17 & 45.83 & 61.11 & 31.67 & 29.10 & 43.34 \\
    \;\; + CB++(shape)    & 70.00 & 46.00 & 38.00 & 38.67 & 57.14 & 54.29 & 54.17 & 45.83 & 61.11 & 33.33 & 29.10 & 44.08 \\
    \;\; + CB++(spatial)   & 86.67 & 39.33 & 32.67 & 39.00 & 44.76 & 49.52 & 58.33 & 44.44 & 44.44 & 21.67 & 27.12 & 41.37 \\
    \;\; + CB++(texture)   & 86.67 & 39.67 & 36.33 & 43.67 & 50.48 & \textbf{66.67} & 58.33 & 52.78 & 50.00 & 21.67 & 31.61 & 43.17 \\
    \;\; + CompAlign & \textbf{90.00} & \textbf{56.00} & \textbf{42.33} & \textbf{51.33} & \textbf{57.14} & 60.00 & \textbf{68.75} & \textbf{54.17} & \textbf{61.11} & \textbf{35.00} & \textbf{41.80} & \textbf{53.08} \\
    \midrule
    \bottomrule
    \end{tabular}
    \vspace{0.2em}
    \captionof{table}{\label{tab:strat-results-full}
    Full experiment results on CompAlign-ed diffusion models, base models, and all checkpoints in ~\citet{10847875}'s work.
    }
\end{table}

\section{Additional Analysis: Are Closed-Source T2I Models Strong Enough?}
From both quantitative and qualitative results reported, it looks like recent advanced closed-source T2I models like DALL-E 3 already achieve near-perfect accuracy on compositional T2I generation.
However, we argue that \textbf{the challenge of compositionally remains far from being solved}.
In this section, we present an interesting observation on how strong T2I models might ``hack'' their way in achieving high compositionally scores, while their outputs still suffer from issues such as naturalness of generated scenes.

\begin{figure}[t]
\includegraphics[width=1.0\textwidth]{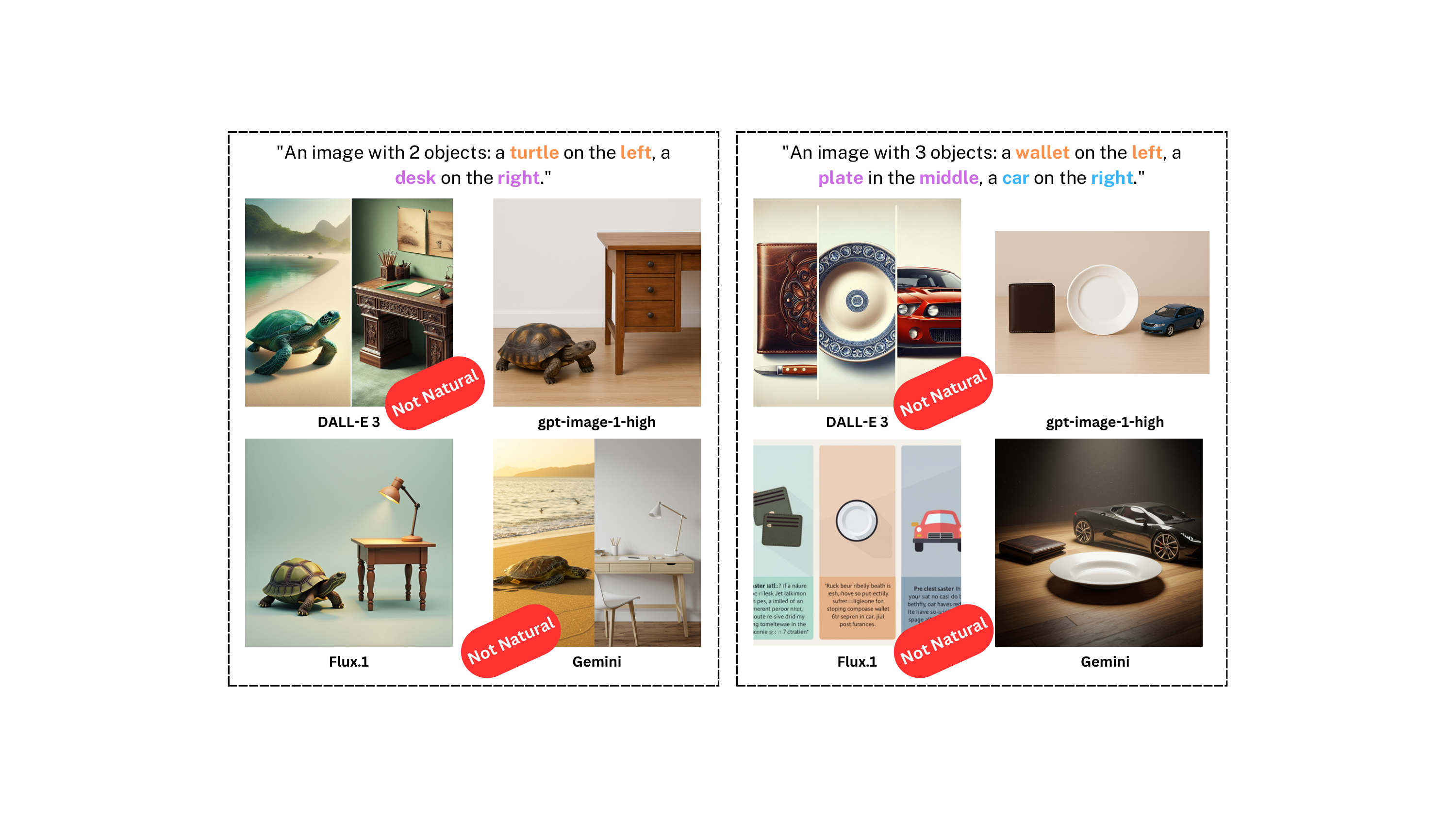}
    \caption{\label{fig:naturalness} Naturalness.}
\end{figure}

Figure \ref{fig:naturalness} demonstrates cases where such pitfall occurs. 
Even for simple compositional T2I prompts involving 2 objects---for instance, a turtle on the left and a desk on the right--- a few numerically ``strong'' T2I models tend to depict the objects in separate scenes, almost looking like 2 separate images collaged together.
While this might ``hack'' the accuracy score in automated compositionally evaluation metrics, the problem with naturalness in generated images will greatly hinder effective downstream applications, e.g. design.

We utilize an MLLM (gpt-4o) to judge the naturalness of generated images for 5 best-performing T2I models on our CompAlign benchmark, by using the prompt in Table \ref{tab:prompt-naturalness}. 
We report the percentage of images judged as ``natural'' by the MLLM in Table \ref{tab:naturalness-results}.
It is easy to observe that even OpenAI's image-1-high model, which achieves near-perfect compositional accuracy, still suffer from naturalness issues.
This case study highlights a critical limitation for T2I models, pointing towards a potential new research direction in T2I evaluation.

\begin{table}[h]
    \centering
    \begin{tabular}{p{0.8\textwidth}}
    \toprule
    \midrule
    \textbf{Naturalness Evaluation Prompt} \\
    \midrule
     'Given an image and its generation prompt with a list of depiction subjects, answer the following question: does the entities in the image appear together in a natural, harmonic scene? If each entity appear as a separate image or in a separate scene, answer no. Only answer yes or no to this question. Prompt: \{\}. Your answer: ' \\
     \bottomrule
    \end{tabular}
    \vspace{1em}
    \caption{Prompt for naturalness evaluation.}
    \label{tab:prompt-naturalness}
\end{table}

\begin{table}[h]
\centering
\begin{tabular}{lc}
\toprule
\textbf{Model} & \textbf{Naturalness Score} \\
\midrule
DALL-E 3 & 40.00 \\
image-1-high & 72.22 \\
SD3.5 & 56.67 \\
Flux.1 & 57.78 \\
Gemini & 66.28 \\
\bottomrule
\end{tabular}
\vspace{1em}
\caption{\label{tab:naturalness-results}Naturalness scores of different models.}
\end{table}

\section{Limitations}
We identify some limitations of our study. 
First, due to cost and resource constraints, we were not able to further extend our experiments to larger scales.
Future works should be devoted to comprehensively evaluating  compositionally of T2I models
Second, experiments in this study incorporate multimodal large language models (MLLMs) that were pre-trained on a wide range of text from the internet and have been shown to learn or amplify biases from the data used.
Therefore, it is possible that certain bias was manifested during experiments.
However, our task is mostly objective and intrinsically neutral to such biases, and we adopted clear prompting instructions to prevent bias from ambiguity.
We encourage future works to also consider this factor in their research, so as to draw reliable and trustworthy research conclusions.

\end{document}